\documentclass[conference,compsoc]{IEEEtran}
\IEEEoverridecommandlockouts
\usepackage{styles/style}
\usepackage{cite}
\usepackage{amsmath,amssymb,amsfonts}
\usepackage{graphicx}
\usepackage{textcomp}
\usepackage{xcolor}
\usepackage{hyperref}
\usepackage{xspace}
\usepackage{stackengine}
\usepackage{float}
\usepackage{bbm}
\usepackage{dsfont}
\usepackage{algorithm}
\usepackage{algorithmic}
\usepackage{enumitem}
\usepackage{booktabs}
\usepackage{footnote}
\usepackage{color, colortbl}
\usepackage{enumitem}
\usepackage{rotating}
\usepackage{hyperref}
\usepackage{dblfloatfix}
\usepackage{url,amsmath,setspace,amssymb}
\usepackage{graphicx}
\usepackage{caption}
\usepackage{subcaption}
\usepackage{multirow}
\usepackage{makecell}
\usepackage{bbold}
\usepackage{cite}

\newcommand{\algname}{MP-Boost\xspace}

\definecolor{commentcolor}{rgb}{0.27, 0.27, 0.33}

\def\BibTeX{{\rm B\kern-.05em{\sc i\kern-.025em b}\kern-.08em
		T\kern-.1667em\lower.7ex\hbox{E}\kern-.125emX}}
\begin{document}
	
	\title{\algname: Minipatch Boosting via Adaptive Feature and Observation Sampling\\
	}
	
	\author{Mohammad Taha Toghani, Genevera I. Allen\thanks{Mohammad Taha Toghani is with the Department of Electrical and Computer Engineering, Rice University, Houston, TX 77005 USA.(email:\href{mailto:mttoghani@rice.edu}{mttoghani@rice.edu}). Genevera I. Allen is with the Departments of Electrical and Computer Engineering, Statistics, and Computer Science, Rice University, Houston, TX 77005 USA, and Jan and Dan Duncan Neurological Research Institute, Baylor College of Medicine, Houston, TX 77030 USA (email:\href{mailto:gallen@rice.edu}{gallen@rice.edu}). The authors acknowledge support from NSF DMS-1554821, NSF NeuroNex-1707400, and NIH 1R01GM140468.}}

	\maketitle
	
	\begin{abstract}
		Boosting methods are among the best general-purpose and off-the-shelf machine learning approaches, gaining widespread popularity.  In this paper, we seek to develop a boosting method that yields comparable accuracy to popular AdaBoost and gradient boosting methods, yet is faster computationally and whose solution is more interpretable.  We achieve this by developing \algname, an algorithm loosely based on AdaBoost that learns by adaptively selecting small subsets of instances and features, or what we term \emph{minipatches} (MP), at each iteration.  By sequentially learning on tiny subsets of the data, our approach is computationally faster than other classic boosting algorithms.  Also as it progresses, \algname adaptively learns a probability distribution on the features and instances that upweight the most important features and challenging instances, hence adaptively selecting the most relevant minipatches for learning.  These learned probability distributions also aid in interpretation of our method.  We empirically demonstrate the interpretability, comparative accuracy, and computational time of our approach on a variety of binary classification tasks.
		
	\end{abstract}
	
	
	\section{Introduction}\label{sec:introduction}
	Boosting algorithms adaptively learn a series of weak learners that overall yield often state-of-the-art predictive accuracy, but are computationally slow.  Huge datasets with many instances and features (e.g., data recorded from sensors, texts, and images) highlight this "slow learning" behavior.  For example, AdaBoost~\cite{freund1997decision} and gradient boosting ~\cite{friedman2001greedy} suffer from a slow training speed since they try to achieve a proper performance neglecting the size of the data.  Further, most boosting methods are "black-box", meaning that their interpretation\cite{lipton2018mythos} lacks transparency and simplicity.  Identification of features and examples with the most impact helps to obtain a more interpretable model~\cite{biau2016random, benard2020interpretable}.  In this paper, our goal is to develop an AdaBoost-based algorithm that learns faster computationally and also yields interpretable solutions.
	
\begin{figure}[t]
		\centerline{\includegraphics[width=75mm]{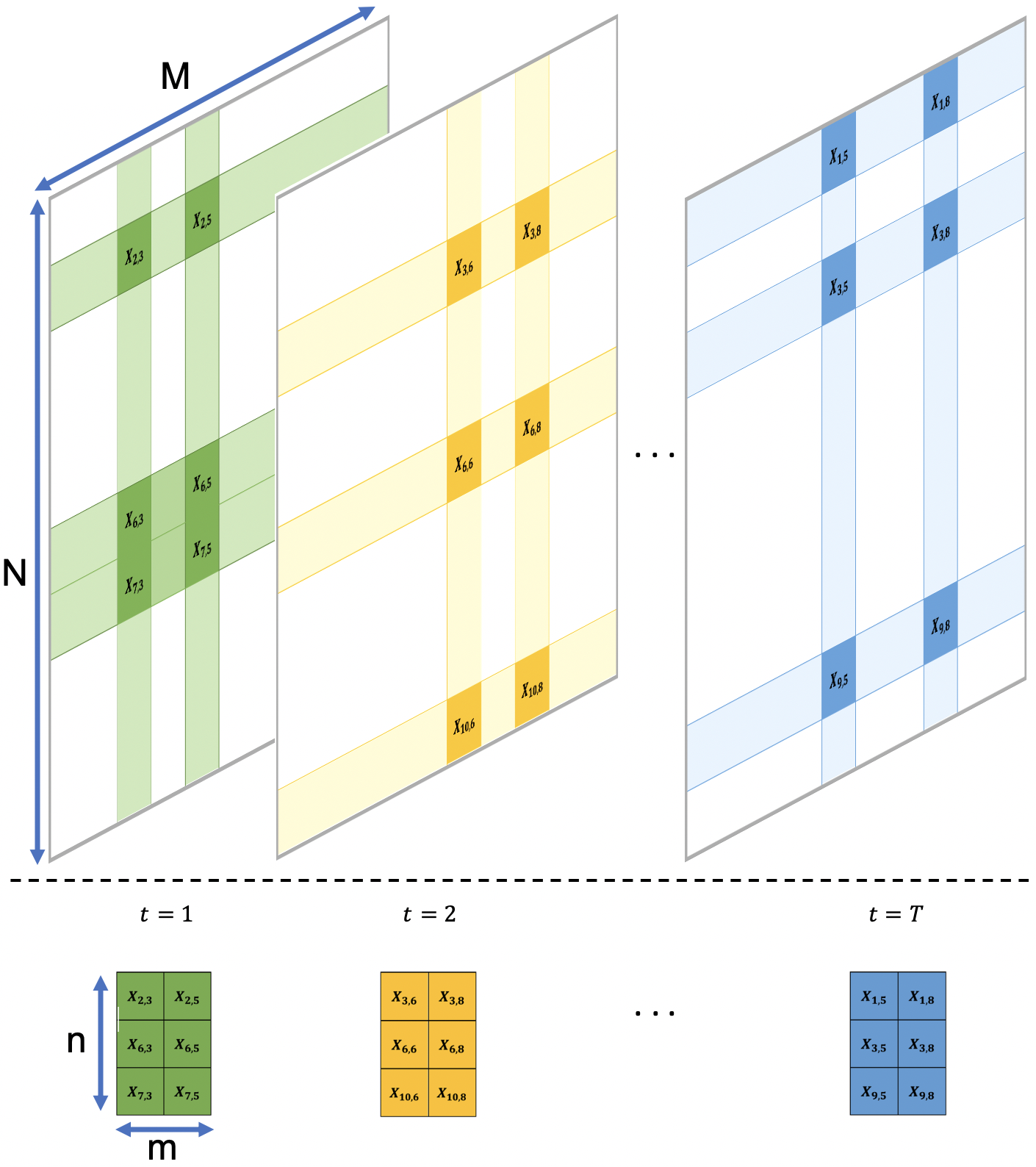}}
		\caption{\textbf{Minipatch Learning}- Selecting random minipatches of the dataset during the training of an ensemble algorithm.}
		\label{fig:minipatch}
	\end{figure}

	We propose to achieve this by adaptively sampling tiny subsets of both instances and features simultaneously, something we refer to as a \emph{minipatch learning} (Fig.~\ref{fig:minipatch}).  Subsampling is widely used in machine learning and has been shown to have both computational and predictive advantages.  For instance, bagging~\cite{breiman1996bagging} uses the bootstrap technique~\cite{Efron1993AnIT} to reduce the dependency of weak learners to the training data.  Random forest~\cite{breiman2001random}, as a specific type of bagging, reduces the number of efficient features in each weak learner as well.  The computational advantages of the ensemble of uniform random minipatches have been investigated in~\cite{louppe2012ensembles}.
	
	There is preliminary evidence that uniform minipatch selection yields implicit regularization~\cite{srivastava2014dropout, lejeune2019implicit}; however, dropout is not precisely what we refer to as minipatch learning.  In fact, the combination of minibatch selection (i.e., stochastic optimization) and dropout in the first layer lies in the category of minipatch learning algorithms. Minipatch sampling not only can accelerate ensemble algorithms, but it can also implicitly regularize them, i.e., reduce the prediction's sensitivity to individual instances or features.
	
	Subsampling can speed up iterative algorithms; however, uniform sampling can appropriately be modified with an adaptive procedure as the importance and difficulty of the observations and features vary.  To give an example, in classification problems, observations closer to the decision boundary play the critical role in the ultimate model~\cite{cortes1995support}.  Moreover, we can use adaptive sampling to interpret the final results.  For instance, SIRUS algorithm~\cite{benard2019sirus,benard2020interpretable} suggests how to leverage the frequency of splits in random forest trees to generate an interpretable model with fewer splits.  Our \algname algorithm will incorporate the advantages of adaptivity in order to learn distributions on the observations and features.
	
	More closely related to our work, several have proposed to employ sampling schemes on the top of AdaBoost.  Works along this line usually target either subsampling the features or observations adaptively.  Exploiting bandit algorithms, adaptive feature selection methods~\cite{busa2009bandit,busa2010fast} have been suggested to reduce the number of effective features accessed by each weak learner.  Similarly, Tasting~\cite{dubout2011tasting} and Laminating~\cite{dubout2011boosting,dubout2014adaptive} propose score-based feature selection algorithms in each iteration of AdaBoost.  They subsample features in each iteration adaptively but either utilize all instances to train weak learners or subsample them uniformly.
	
	On the other hand, algorithms like MadaBoost~\cite{domingo2000madaboost} and FilterBoost~\cite{bradley2008filterboost} suggest adaptive oracle subsampling of the observations.  However, they do not subsample features and the size of selected samples increases per iteration, thus get slower during their progress.  In contrast, the algorithm that we will propose exploits the effect of minipatch selection; in this way, it will take the importance of both the observations and features into account. 
	
	A series of algorithms have been proposed to reduce the computational complexity of gradient boosting.  For example, stochastic gradient boosting (SGB)~\cite{friedman2002stochastic} suggests subsampling the observations randomly.  XGBoost~\cite{chen2016xgboost} by introducing a novel structure for decision trees, a fast training algorithm, as well as several other modifications such as features subsampling techniques through a pre-sorting and histogram-based algorithm, extensively optimizes the computational complexity of gradient boosting.
	
	In this regard, LightGBM~\cite{ke2017lightgbm} and CatBoost~\cite{dorogush2018catboost} subsample observations adaptively proportionate to their gradient values and use an algorithm called exclusive feature bundling (EFB) to reduce the number of effective features by categorizing them.  In contrast, \algname will be designed to learn a probability distribution on features gradually during its progress instead of categorizing them initially.  The minimal variance sampling (MVS)~\cite{ibragimov2019minimal} algorithm is proposed to select the observations according to their gradient values provably with a minimal variance; however, it lacks subsampling over the features. Note that \algname will be designed based on AdaBoost; thus, its adaptive observation selection will entirely be different from algorithms that are designed based on gradient boosting.
	
	The remainder of the paper is organized as follows. In Section~\ref{sec:main}, stating the problem formulation, we present \algname. In Section~\ref{sec:experiments}, we investigate the efficacy of the adaptive subsampling, interpretability of the proposed algorithm, and compare the generalization accuracy of \algname with that of AdaBoost, gradient boosting, and random forest. We end with discussing and concluding remarks in Section~\ref{sec:discussion}.
	
	\section{\algname}\label{sec:main}
	Our goal is to develop a boosting algorithm utilizing adaptive sampling of features and observations that enhances both scalability and interpretability.  
	
	\subsection{\algname Algorithm}
	We begin by focusing on binary classification tasks.  Let our data be $X \in \bbR^{N \times M}$ for $N$ observations or instances and $M$ features; for each instance, we observe a label, $(\vx_i,y_i)$ with $y_i\in\{-1,+1\}$.  We seek to learn a classifier $y_i = \sgn(F(\vx_i))$.
	
	To achieve this, we propose an adaptive sampling based version of boosting inspired by AdaBoost.  Our method relies on learning a weak learner from a tiny subset of observations and features at each iteration.  We call this tiny subset a \emph{minipatch}, termed based on the use of "patches" in image processing, and minibatches as small subsamples of observations commonly used in machine learning.  Our approach is to take an ensemble of minipatches, or minipatch learning as shown in Figure~\ref{fig:minipatch}, where each minipatch is sampled adaptively.  Formally, we define a minipatch by $(\mX_{{\textstyle\mathstrut}\mcR,\mcC}, \vy_{{\textstyle\mathstrut}\mcR})$, where $\mcR$ is a subset of observations with size $n$, and $\mcC$ is a subset of features with size $m$.  By learning weak learners on tiny subsets or minipatches, our boosting algorithm will have major computational advantages for large $N$ and/or large $M$ datasets.     
	
	We define $\mcH$ to be the \emph{class of weak learners}~\cite{mohri2018foundations}, where each $h\in \mcH$ is a function $h:\bbR^m\rightarrow\{-1,+1\}$.  Our algorithm is generic to the type of weak learners, which can be either simple or expressive. However, we select decision trees~\cite{breiman1984classification} as the default weak learner in \algname.  We consider both depth-$k$ trees as well as saturated trees that are split until each terminal leaf consists of samples from the same class.   
	
	The core of our algorithm uses adaptive sampling of observations to achieve the adaptive slow learning properties~\cite{wyner2017explaining} of the AdaBoost algorithm.  Similar to \cite{ibragimov2019minimal} for gradient boosting, \algname subsamples observations according to an adaptive probability distribution. Let $\vlp$ be the probability distribution on observations (i.e., $\sum_{i=1}^{N} p_i = 1$) and initially set $\vlp$ to be uniform ($\mU_{[N]}$).  We define $\mathrm{Sample}(N,n,\vlp)$ as sampling a subset of $[N]$ of size $n$ according to the probability distribution $\vlp$ without replacement.
	
	Let $F:\bbR^M\to \bbR$ be the \emph{ensemble function}.
	Our algorithm selects a minipatch, trains a proper weak learner on it, and computes the summation of weak learners, $F(\vx_i)^\ptp = \sum_{k=1}^t h^{(k)}\big({(\vx_i)}_{{\textstyle\mathstrut}\mcC^{(k)}}\big)$.  Misclassified samples are more difficult to be learned, so we need to increase their probabilities to be sampled more frequently.  Let $\mcL: \bbR \times \bbR \to \bbR^{+}$ be a function that measures the similarity between the ensemble outputs and labels, i.e., positive $yF$ yields smaller $\mcL(y,F)$ and vice versa.  \algname assigns a probability proportional to $\mcL(y_i,F(\vx_i)$ to the $i^\text{th}$ observation.  Table~\ref{tab:probability-update} shows the choices for function $\mcL$ inspired by the weighting function in AdaBoost and LogitBoost~\cite{friedman2000additive}.  Note that "Soft functions" are sensitive to the ensemble output value, while "Hard functions" merely care about its sign.
	
	Full-batch boosting algorithms reweight all of the observations and train a new weak learner on their weighted average in each iteration~\cite{freund1997decision,friedman2000additive,friedman2001greedy}.  In contrast, stochastic algorithms use each sample's frequency to take the effect of its weight into account\cite{domingo2000madaboost,bradley2008filterboost,ibragimov2019minimal}.  We update the probability of the observations similar to FilterBoost~\cite{bradley2008filterboost}. However, FilterBoost increases $n$ during its progress and uses negative sampling, and is thus slower than ours.
	
	\begin{table}[t]
		\caption{$\boldsymbol{\mcL(y_i,F(\vx_i))}$ \textbf{choices for weighting samples}- $\mcL$ is a decreasing function with respect to $y_i\times F(\vx_i)$. Hard functions depend on the sign of $y_iF(\vx_i)$ that makes \algname less sensitive to outlier samples.}
		\centering
		\begin{tabular}{ccc}
			\toprule
			&\textbf{Exponential}&\textbf{Logistic}\\
			\cmidrule(lr){2-3}
			\textbf{Soft}&$\exp(-y_{i} F({\vx_i}))$&$\frac{1}{1+\exp(y_{i} F({\vx_i}))}$\\[2mm]
			\textbf{Hard}&$\exp(-y_{i} \sign(F({\vx_i})))$&$\frac{1}{1+\exp(y_{i} \sign(F({\vx_i})))}$\\
			\bottomrule
		\end{tabular}
		\label{tab:probability-update}
	\end{table}
	
	Another major goal is to increase the interpretability of boosting approaches.  To accomplish this, we also propose to adaptively select features that are effective for learning.  Similar to $\vlp$, let $\vlq$ be the probability distribution on features.  Our algorithm requires a criterion to compute the importance of the selected $m$ features based on the structure of $\mcH$.  There exist several choices for computing features importance based on $\mcH$. Some of these inspection techniques are model agnostic, hence proper for \algname to incorporate weak learners from different classes.  For example, the permutation importance method~\cite{breiman2001random,altmann2010permutation} shuffling each feature infers its importance according to the difference in the prediction score.
	
	Nevertheless, specific metrics like \emph{impurity reduction score}~\cite{nembrini2018revival} are defined for decision trees.  We utilize this quantity to define our probability distribution over the features.  Let $\mcI^{h}$ denote the normalized feature importance vector for a weak learner $h$, wherein each entry determines the relative importance of the corresponding feature compared to other features in the minipatch.  In each iteration, $\vlq$ is updated through computing the weighted average of $\vlq$ and $\mcI$ according to a momentum $\mu$.  The hyperparameter $\mu\in(0,1)$ determines the ratio of exploration vs. exploitation.  \algname only modifies the probability of features inside the minipatch, in each iteration.
	
	\begin{algorithm}[t]
		\caption{\algname}
		\textbf{\algname} $(\mX, \vy, n, m, \mu)$\\
		\textbf{\color{commentcolor}Initialization ($t=0$):}\\
		$\vlp^{(1)} = \mU_{[N]}\quad$ \textcolor{commentcolor}{// observation probabilities} \\
		$\vlq^{(1)} = \mU_{[M]}\quad$ \textcolor{commentcolor}{// feature probabilities}\\
		$F^{(1)}(\vx_i)=0, \quad \forall i\in [N]\quad$ \textcolor{commentcolor}{// ensemble output}\\
		$G^{(1)}(\vx_i)=0, \quad \forall i\in [N]\quad$ \textcolor{commentcolor}{// out-of-patch output}\\
		\textbf{while} $\mathrm{Stopping-Criterion}(\oop^\ptp)$ not met \textbf{do} $t\leftarrow t+1$
		\begin{enumerate}
			\item \textcolor{blue}{\textbf{Sample a minipatch:}}\label{step:sample-minipatch}
			\begin{enumerate}
				\item $\mcR^\ptp=\mathrm{Sample}(N,n,\vlp^\ptp) \quad$ \textcolor{commentcolor}{// select $n$ instances}\label{step:sample-observations}
				\item $\mcC^\ptp=\mathrm{Sample}(M,m,\vlq^\ptp) \quad$ \textcolor{commentcolor}{// select $m$ features}\label{step:sample-features}
				\item $\big(\mX^\ptp,\vy^\ptp\big)=\big(\mX_{{\textstyle\mathstrut}\mcR^\ptp,\mcC^\ptp},\vy_{{\textstyle\mathstrut}\mcR^\ptp}\big) \quad$ \textcolor{commentcolor}{// minipatch}\label{step:minipatch}
			\end{enumerate}
			\item \textcolor{blue}{\textbf{Train a weak learner on the minipatch:}}\label{step:train-weak-learner-main}
			\begin{enumerate}
				\item $h^\ptp \in \mcH$: weak learner trained on $\mX^\ptp, \vy^\ptp$\label{step:train-weak-learner}
			\end{enumerate}
			\item \textcolor{blue}{\textbf{Update outputs:}}
			\begin{enumerate}\label{step:update-outputs-main}
				\item $F^\ptp(\vx_i) = F^{(t-1)}(\vx_i) + h^\ptp\big({(\vx_i)}_{{\textstyle\mathstrut}\mcC^\ptp}\big), \quad \forall i\in [N]$\label{step:update-outputs}
			\end{enumerate}
			\item \textcolor{blue}{\textbf{Update probability distributions:}}\label{step:update-distributions}
			\begin{enumerate}
				\item $p^{(t+1)}_{i} = \frac{\mcL(y_i,F^\ptp(\vx_i))}{\sum_{k=1}^{N}\mcL(y_k,F^\ptp(\vx_k))}, \quad \forall i\in [N] \quad$\label{step:update-observations-probability}
				
				\vspace{3mm}
				\item$q^{(t+1)}_{j} = 
				(1-\mu) q^\ptp_{j} + \mu r\mcI^{h^\ptp}_j, \quad j\in \mcC^\ptp$\\
				where, $r = \displaystyle\sum_{j \in \mcC^\ptp} q_j^\ptp$\label{step:update-features-probability}
			\end{enumerate}
			\item \textcolor{blue}{\textbf{Out-of-Patch Accuracy:}}\label{step:oop}
			\begin{enumerate}
				\item $G^\ptp(\vx_i) = G^{(t-1)}(\vx_i) +  h^\ptp\big({(\vx_i)}_{\mcC^\ptp}\big), \quad \forall i\notin \mcR^\ptp$\label{step:update-oop-outputs}
				
				\vspace{1.5mm}
				\item $\oop^\ptp= \frac{1}{N}\sum_{i=1}^{N} \mathds{1}_{\{\sign(G^\ptp(\vx_i))=y_i\}}$\label{step:compute-oop-accuracy}
			\end{enumerate}
		\end{enumerate}
		\textbf{end while}\\
		\textbf{Return} $\sign(F^{(T)}),\vlp^{(T)},\vlq^{(T)}$
		
		\label{alg:main}
	\end{algorithm}

	Finally, many boosting algorithms are designed to run for a fixed number of iterations\cite{friedman2001greedy} or use a validation criterion~\cite{domingo2000madaboost, bradley2008filterboost, Meijer2016RegularizingAW} in order to determine when to stop.  Internal validation approaches often have better performance and are computationally much faster.  For instance, consider the out-of-bag criterion~\cite{breiman1996out} in bagging and random forest that uses internal validation properties without incurring any additional computational cost.  Similarly to bagging, our \algname has access to out-of-patch instances, which we can use for internal validation.  Therefore, for each sample $i\in[N]$, we accumulate the output of weak learners that do not have it in their minipatch.  Thus, we define \emph{out-of-patch output} to be a function $G:\bbR^M\to \bbR$ as follows:
	\begin{align}
		G^\ptp(\vx_i) = \sum_{k=1}^{t} h^{(k)}\big({(\vx_i)}_{{\textstyle\mathstrut}\mcC^{(k)}}\big)\mathbb{1}_{\{i\notin\mcR^{(k)}\}}
	\end{align}
	for an arbitrary $\vx_i$. Accordingly, the out-of-patch accuracy, $\oop$, can easily be quantified.
	
	Out-of-patch accuracy is a conservative estimate of the test accuracy. Hence it can assist \algname to track the progress of the generalization (test) performance internally and decide at which iteration to stop.  In a nutshell, observing the $\oop$ value, the algorithm finds where it is saturated.  Algorithm~\ref{alg:stop} (in Appendix~\ref{sec:stopping-criterion}) is a heuristic algorithm that takes $\oop^\ptp$, compares it with its previous values, and finally decides when it becomes saturated.  In fact, the stopping algorithm follows a general rule; if the current value of $\oop$ increases with some margin, then the algorithm needs more time to improve; otherwise, the generalization performance is saturated.
	
	We put all of this together in a summary of our \algname algorithm in Algorithm~\ref{alg:main}.  Notice here that selecting minipatches (step \ref{step:sample-minipatch}) reduces the computational complexity imposed per iteration, thus improves the scalability.  Other variations of AdaBoost usually subsample either features or observations while ours exploits both.  Therefore, in addition to the predictive model $F$, our algorithm learns probability distributions $\vlp$ and $\vlq$ that express the importance of observations and features, respectively. Since learning $\vlp,\vlq$ is a part of the iterative procedure, it does not incur an extra computational cost.  In addition, \algname exploits an internal validation that yields an automatic stopping criterion when the algorithm ceases to learn.  Hence, steps (\ref{step:update-distributions}) and (\ref{step:oop}) of Algorithm~\ref{alg:main} highlight the main differences of \algname with other sampling-based boosting algorithms.

	\subsection{Hyperparameter Tuning}
	The minipatch size is a crucial hyperparameter of our algorithm.  Large $n$ or $m$ slows down weak learners' training but is more likely to yield better performance. Note that $m$ must be large enough to provide a wide range of features for comparison and update features probability properly.  Similarly, $n$ must be large enough such that each minipatch represents a meaningful subset of the observations.  On the other hand, small $n$ results in a $\oop$ value akin to the generalization accuracy.  Selecting around ten percent of the observations and features seems to be a proper choice for our algorithm, as evidenced in our empirical studies.  Additionally, our studies reveal that the results are fairly robust to small changes in $n$ and $m$.  $\mu$ is the other important hyperparameter in our algorithm where a moderate value for it (e.g., $\mu=0.5$) strikes a balance between exploration and exploitation.  While our default hyperparameter settings seem to perform well and are robust in a variety of settings (see Section~\ref{sec:experiments}), one could always select these in a data-driven manner using our $\oop$ criterion as well.
	
	\subsection{Extensions}
	Our proposed algorithm is initially developed for the binary classification problem.  Here, we discuss its extension to \emph{regression} and \emph{multiclass classification} problems.  First, for multiclass classification, algorithms like AdaBoost.M2, AdaBoost.MH, and AdaBoost.OC~\cite{schapire1997using,chengsheng2017adaboost} are proposed as multiclass extensions of AdaBoost.  We can incorporate similar modifications to extend \algname as well.  Further,~\cite{drucker1997improving,solomatine2004adaboost} have suggested regression extensions like AdaBoost.R2 and AdaBoost.RT to the vanilla AdaBoost.  Obviously, for regression, we will have to change our observation probability function and the out-of-patch accuracy, thus we can employ techniques in~\cite{breiman1996out} to address the regression problem.  All these can be further extensions to our approach.

	\section{Experiments}\label{sec:experiments}
	We begin by using an illustrative case study to show how our method works and how it aids interpretability.  Next, we compare our algorithm to other popular boosting and tree-based methods, focusing on accuracy and scalability.
	
	\begin{figure}[!b]
		\begin{subfigure}[b]{\linewidth}
			\centering
			\centerline{\hspace{0.9em}\includegraphics[width=\linewidth]{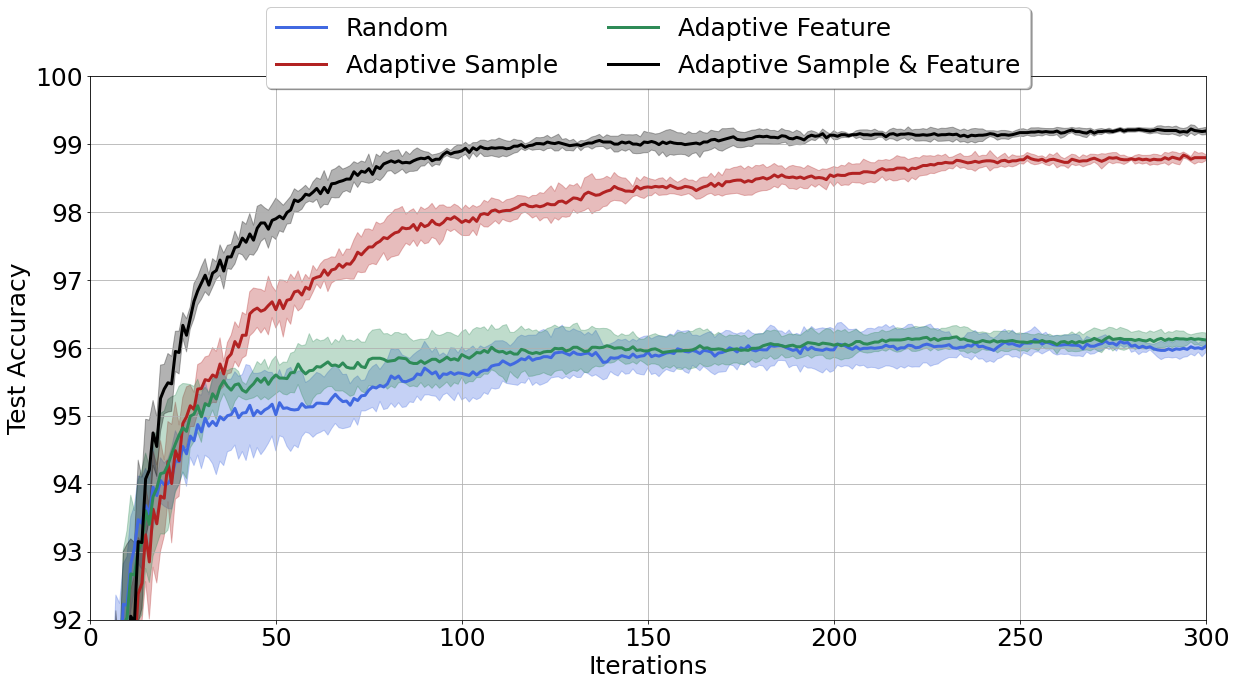}}
			\caption{}
			\label{fig:adaptivity}
		\end{subfigure}
		\begin{subfigure}[b]{\linewidth}
			\centering
			\centerline{\includegraphics[width=\linewidth]{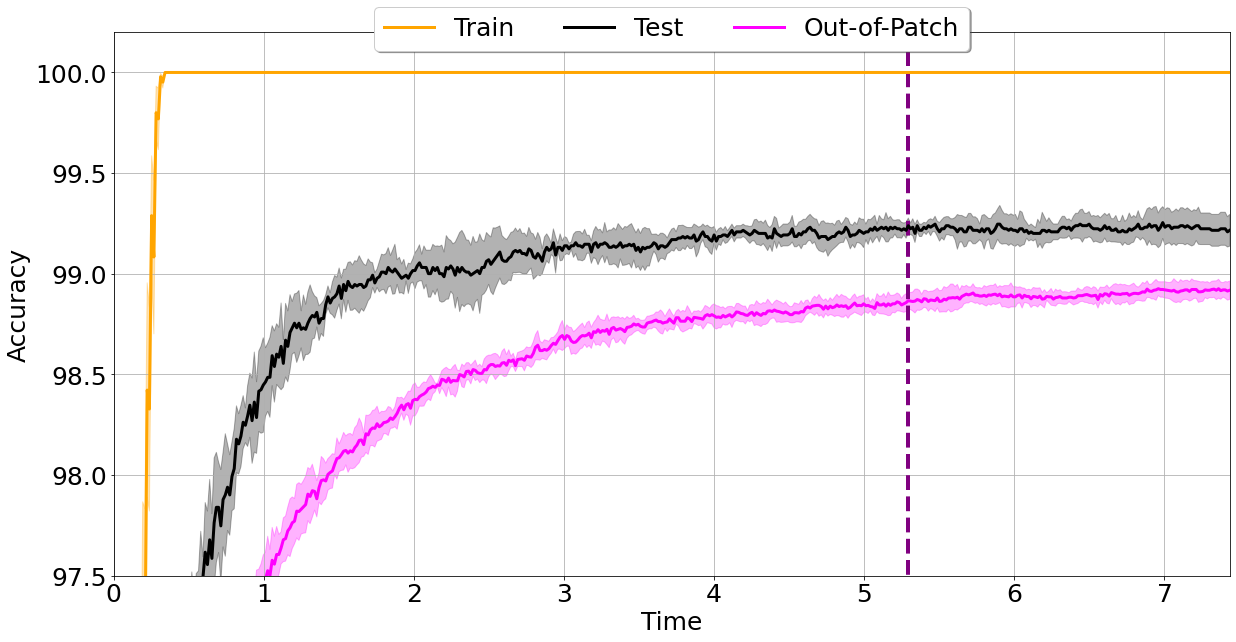}}
			\caption{}
			\label{fig:oop}
		\end{subfigure}
		\caption{(a)~Effect of \algname adaptive sampling on MNIST ($3$ vs. $8$) test data. (b)~Train, test, and out-of-patch accuracy of MNIST ($3$ vs. $8$) using \algname. The dashed line shows the stopping time based on Algorithm\ref{alg:stop}.}
		\label{fig:mix-adaptivity-oop}
	\end{figure}

	\subsection{Illustrative Case Study}
	We use a series of experiments to demonstrate how our algorithm works and show how to interpret the results.  Specifically, we ask: how does adaptive sampling boost the performance; how does the out-of-patch accuracy relate to test accuracy and yield a data-driven stopping criterion; how do we interpret the results via the final value of $\vlp$ and $\vlq$?
	
	To answer these questions, we focus our investigations on an explicable binary classification task: detecting digit $3$ versus $8$ in MNIST~\cite{lecun1998mnist}.  This dataset includes handwritten digits as images of size $28\times 28$.  The training data is huge ($N>10000$) and high-dimensional ($M=784$).  We use cross-validation to tune all hyperparameters, yielding  $n=500$, $m=30$, and $\mu=0.5$ as well as the Soft-Logistic function~\ref{tab:probability-update} as $\mcL$.  
	
	\begin{figure*}[!ht]
		\centering
		\begin{subfigure}[b]{0.49\textwidth}
			\centering
			\includegraphics[width=\textwidth]{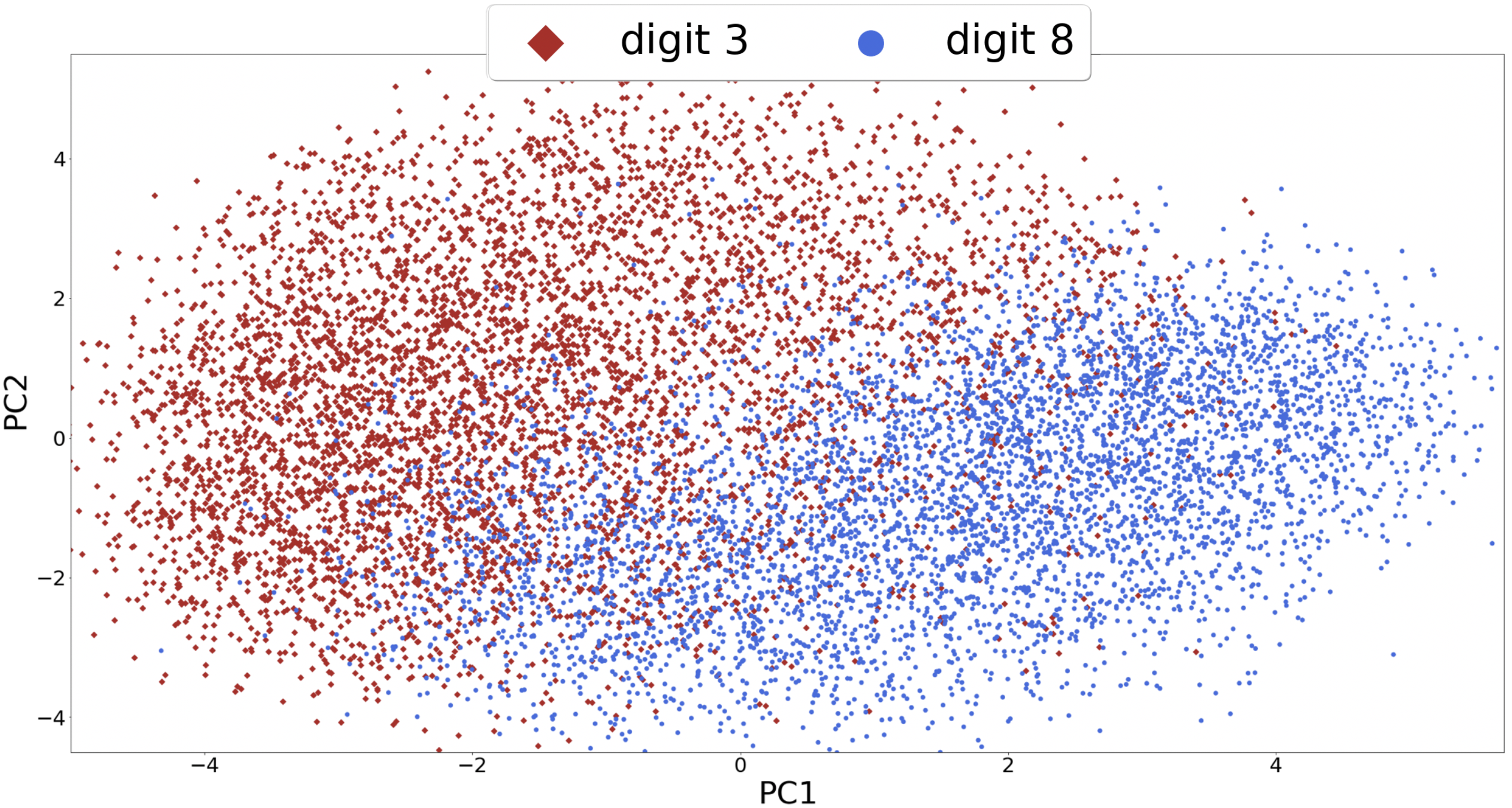}
			\caption{{\small Before training}}    
			\label{fig:s-prob-before}
		\end{subfigure}
		\begin{subfigure}[b]{0.49\textwidth}  
			\centering 
			\includegraphics[width=\textwidth]{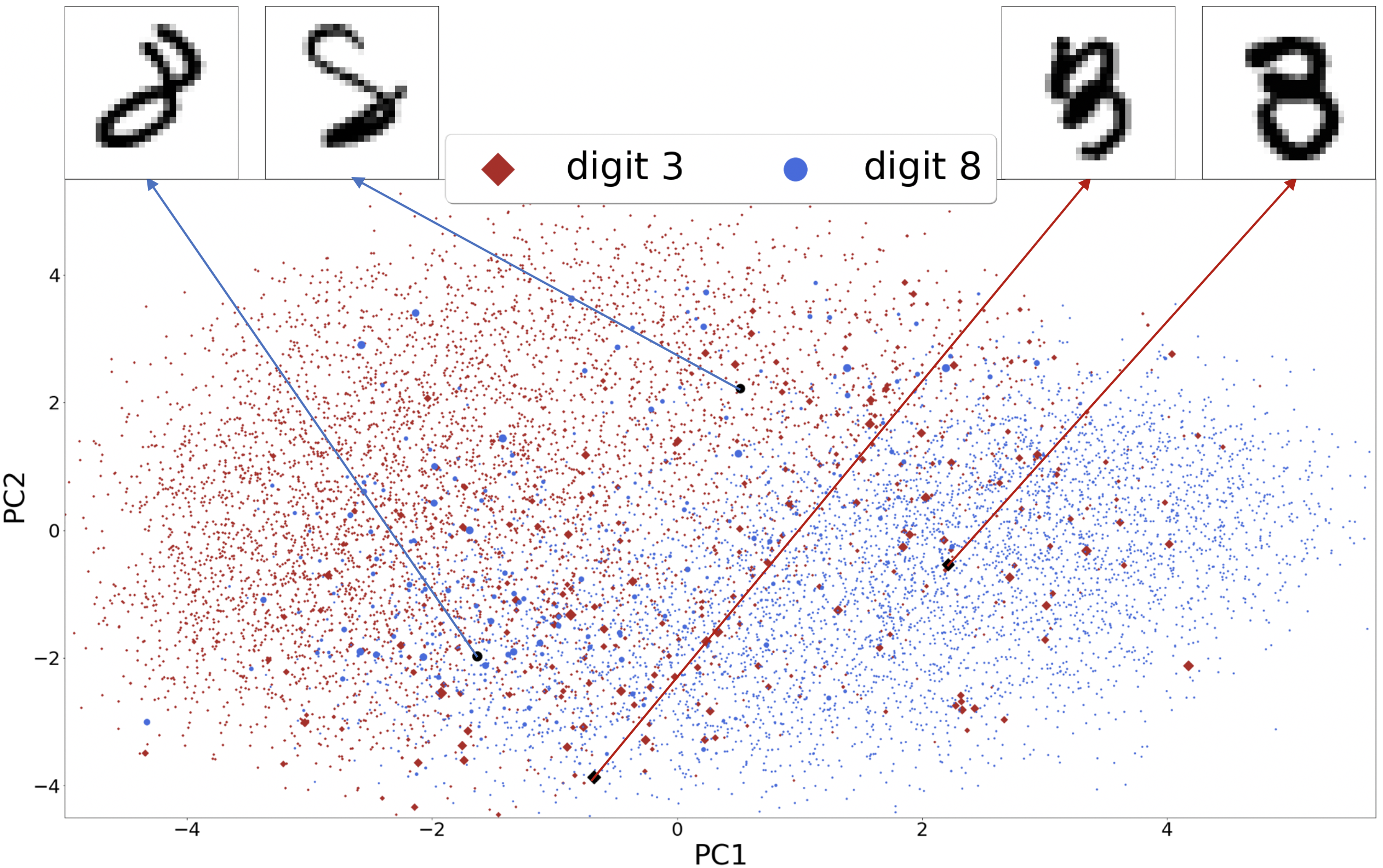}
			\caption{{\small After training}}   
			\label{fig:s-prob-after}
		\end{subfigure}
		\caption
		{\textbf{Probability distribution of MNIST (digit $\mathbf{3}$ vs. $\mathbf{8}$) samples before and after training \algname}- Samples are Projected onto a $2$-dimensional space using PCA. The size of each point (sample) indicates its relative probability. After training, samples close to the boundary have higher probabilities. The displayed samples are visually hard to be learned and have high probabilities.}
		\label{fig:sample-prob}
	\end{figure*}
	
	To measure the effect of adaptive observation and/or feature selection, we train \algname on MNIST($3,8$).  Then, we turn off the adaptive updates for $\vlp$ and replace it with the uniform distribution and resulting random observation sampling.  We do the same for $\vlq$ analogously for the features.  Finally, we turn off adaptive sampling for both features and observations, and repeat each experiment $5$ times. Figure~\ref{fig:adaptivity} shows the superiority of joint adaptive sampling of both observations and features in terms of performance on the test data.  Further, we compare the training, out-of-patch, and test accuracy curves for \algname in Figure~\ref{fig:oop}.  The dashed line indicates the stopping time of our algorithm based on the $\oop$ curve and our stopping heuristic Algorithm~\ref{alg:stop}.  To observe the behavior of the three curves, we let \algname continue after the stopping criterion is satisfied.  As shown in Figure~\ref{fig:oop}, and unlike the training curve, the trend in the out-of-patch curve is similar to that of the test curve. These results demonstrate the power that adaptive sampling of both observations and features brings to boost performance as well as the ability to use the $\oop$ curve to assess algorithm progress.
	
	Next, we illustrate how to use $\vlp$ and $\vlq$ to interpret the observations and features.  First, observations that are difficult to classify are upweighted in $\vlp$.  Hence, we can use $\vlp$ to identify the most challenging samples, yielding a similar type of interpretation commonly employed in support vector machines \cite{cortes1995support}.  To visualize the final value of $\vlp$, we project observations on a two-dimensional space using PCA with sizes of each observation proportional to $\vlp$.  We show this visualization before and after the training in Figure~\ref{fig:sample-prob}.  Further, we highlight a few of the observations with large values of $\vlp$, illustrating that these samples are indeed difficult to distinguish between the two classes.

	\begin{figure*}[!ht]
		\centering
		\begin{subfigure}[b]{0.13\textwidth}
			\centering
			\includegraphics[width=\textwidth]{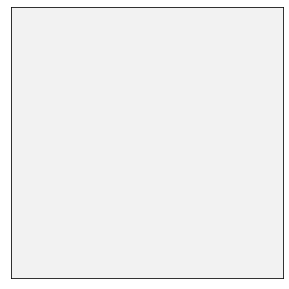}
			\caption{{\small $t=0$}}    
			\label{fig:feature-prob-000}
		\end{subfigure}
		\begin{subfigure}[b]{0.13\textwidth}
			\centering
			\includegraphics[width=\textwidth]{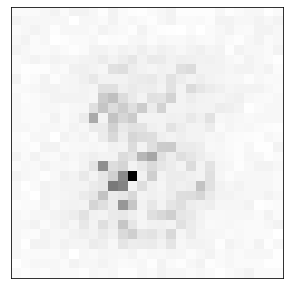}
			\caption{{\small $t=10$}}    
			\label{fig:feature-prob-010}
		\end{subfigure}
		\begin{subfigure}[b]{0.13\textwidth}
			\centering
			\includegraphics[width=\textwidth]{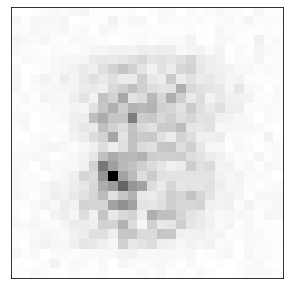}
			\caption{{\small $t=30$}}
			\label{fig:feature-prob-030}
		\end{subfigure}
		\begin{subfigure}[b]{0.13\textwidth}
			\centering
			\includegraphics[width=\textwidth]{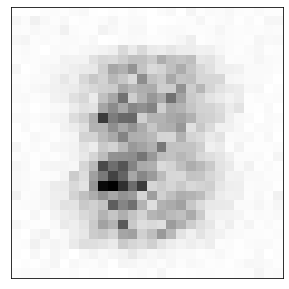}
			\caption{{\small $t=50$}}    
			\label{fig:feature-prob-050}
		\end{subfigure}
		\begin{subfigure}[b]{0.13\textwidth}
			\centering
			\includegraphics[width=\textwidth]{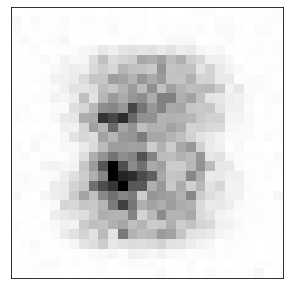}
			\caption{{\small $t=100$}}    
			\label{fig:feature-prob-100}
		\end{subfigure}
		\begin{subfigure}[b]{0.13\textwidth}
			\centering
			\includegraphics[width=\textwidth]{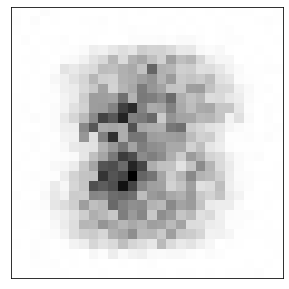}
			\caption{{\small $t=200$}}    
			\label{fig:feature-prob-200}
		\end{subfigure}
		{\vline width 0.5mm}
		\begin{subfigure}[b]{0.13\textwidth}
			\centering
			\includegraphics[width=\textwidth]{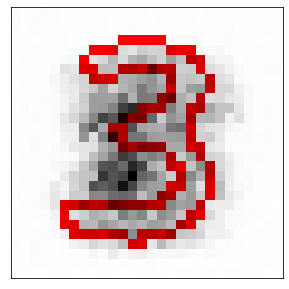}
			\caption{{\small final}}    
			\label{fig:feature-prob-final}
		\end{subfigure}
		\caption
		{\textbf{Probability distribution of MNIST (digit $\mathbf{3}$ vs. $\mathbf{8}$) features during the training of \algname}- The color of each pixel (feature) indicates the relative probability with respect to which the feature is selected. As \algname progresses, the probability of ineffective features (pixels in the background) decreases while that of efficient features (pixels shaping the digits) increases. The final distribution resembles a blurred version of digits $3$ and $8$.} 
		\label{fig:feature-prob}
	\end{figure*}
	
	Finally, we show how to use $\vlq$ to find the most important features, and also illustrate how \algname learns these features probability in Figure~\ref{fig:feature-prob}.  Here, the color of each pixel (feature) is proportional to its probability, with darker pixels indicating the feature has been upweighted. We expect a sparse representation for $\vlq$ which matches with our result.  Moreover, this example clearly shows how to interpret the most relevant features; in Figure~\ref{fig:feature-prob-200}, two regions are darker compared to other pixels corresponding with the complementary area for digit $3$ versus $8$.
	
	\subsection{Comparative Empirical Results}
	Here, we compare the speed and performance of our algorithm with AdaBoost, gradient boosting, and random forest on multiple binary classification tasks.   To this end, we select large real datasets (usually $N>1000$ and $M>100$) from UCI machine learning repository~\cite{Dua:2019}, MNIST~\cite{lecun1998mnist}, and CIFAR-10\cite{krizhevsky2009learning}. Moreover, we use a sparse synthetic dataset of two high-dimensional cones.  The size of datasets is provided in Appendix~\ref{sec:datasets}.
	
	To be fair, we choose the oracle hyperparameters for every method.  To this end, we pick decision trees with different maximum depths ($\text{depth}\in\{1,2,3,4,5,6,7\}$) or depth-saturated trees as weak learners.  Note that we use scikit-learn~\cite{scikit-learn} modules to implement our algorithm and all competitors so that all time-based comparisons are fair.  For our method, we select from the following choices of hyperparameters: $n\in\{50,100,200,500\}$, $m\in\{5,10,15,20,\sqrt{M}\}$, and $\mu\in\{0.1,0.3,0.5,0.7,0.9\}$.
	
	For each dataset, we select the best performance of \algname, versus that of AdaBoost, gradient boosting, and random forest constrained to the runtime of \algname.  Table~\ref{tab:fixed-runtime} shows the best performance of each algorithm within the fixed runtime (\algname training time). Results indicate that \algname achieves a better performance faster than the other three algorithms.  We also provide a more comprehensive comparison in Table~\ref{tab:full-comparison}, where we can see that without any runtime constraint, \algname is much faster with a comparable accuracy across a wide variety of datasets.

	\begin{table*}[!b]
		\caption{Best test accuracy (\%) of \algname, AdaBoost, Random Forest, and Gradient Boosting on binary classification tasks during the runtime of \algname.}
		\centering
		\begin{tabular}{lcccc}
			\toprule
			\textbf{Dataset} & \textbf{\algname} & \textbf{AdaBoost} & \textbf{Random Forest} & \small\textbf{Gradient Boosting}\\
			\midrule
			\small\textbf{Cones} &$\mathbf{100.0 \pm 0.0}$	 &$85.56 \pm 0.16$ & $95.89 \pm 0.97$& $94.01 \pm 1.23$ \\
			\small\textbf{Hill-Valley} &$\mathbf{99.45 \pm 0.19}$	 &$96.28 \pm 1.47$ & $98.76 \pm 0.34$& $96.69 \pm 0.67$ \\
			\small\textbf{Christine} &$\mathbf{74.27 \pm 0.32}$	 &$69.28 \pm 0.09$ & $73.75 \pm 0.04$ & $70.27 \pm 0.13$ \\
			\small\textbf{Jasmine} &$\mathbf{80.03 \pm 0.76}$ &$79.53 \pm 0.01$  & $79.47 \pm 0.55$ & $79.47 \pm 0.08$ \\
			\small\textbf{Philippine} &$\mathbf{71.01 \pm 0.37}$ &$69.64 \pm 0.01$ & $70.07 \pm 0.31$ & $69.98 \pm 0.01$ \\
			\small\textbf{SensIT Vehicle} &$\mathbf{86.23\pm 0.01}$ &$83.69 \pm 0.01$  & $85.98 \pm 0.05$ & $84.24 \pm 0.02$ \\
			\small\textbf{Higgs Boson} &$83.42 \pm 0.06$ &$82.52 \pm 0.01$  & $\mathbf{83.57 \pm 0.08}$ & $81.44 \pm 0.01$ \\
			\small\textbf{MNIST (3,8)} &$\mathbf{99.31 \pm 0.03}$ &$97.49 \pm 0.08$  & $98.84 \pm 0.03$ & $96.81 \pm 0.03$ \\
			\small\textbf{MNIST (O,E)} &$\mathbf{98.15 \pm 0.06}$ &$93.78 \pm 0.01$  & $97.74 \pm 0.04$ & $93.23 \pm 0.01$ \\
			\small\textbf{CIFAR-10 (T,C)} &$\mathbf{73.67 \pm 0.39}$ &$67.65 \pm 0.02$  & $73.08 \pm 0.4$ & $67.81 \pm 0.15$ \\
			\small\textbf{GAS Drift} &$\mathbf{99.76 \pm 0.04}$ &$96.54 \pm 0.02$  & $99.64 \pm 0.08$ & $96.54 \pm 0.02$ \\
			\small\textbf{DNA} &$\mathbf{97.44 \pm 0.07}$ &$96.86 \pm 0.17$  & $\mathbf{97.44 \pm 0.2}$ & $96.81 \pm 0.07$ \\
			\small\textbf{Volkert} &$74.14\pm 0.08$ &$71.54 \pm 0.01$  & $\mathbf{77.95 \pm 0.05}$ & $71.9 \pm 0.04$\\
			\bottomrule
		\end{tabular}
		\label{tab:fixed-runtime}
	\end{table*}

	\section{Discussion}\label{sec:discussion}
	
	In this work, we proposed a new boosting approach using adaptive minipatch learning. We showed that our approach is computationally faster, as well as more interpretable, compared to standard boosting algorithms.  Moreover, we showed that our algorithm outperforms AdaBoost and gradient boosting in a fixed runtime and has a comparable performance without any time constraint.
	
	Our approach would be particularly useful for large data settings where both observations and features are large.  Further, our \algname algorithm can be particularly fitting to datasets with sparse features or noisy features because we learn the important features.  Further, our adaptive observation selection can be used in active learning problems.
	
	We suggested extensions of \algname for multiclass classification and regression problems in this paper.  In future work, we plan on comparing them empirically to existing algorithms.  Other aspects of future work would be to use this minipatch learning scheme in not only AdaBoost-like methods but also in gradient boosting-based methods.  Additionally, one could explore other updating schemes for both the observation and feature probabilities.  Theoretical analysis of \algname would help to interpret the performance of our algorithm and its properties.  Efficient implementations can improve the speed of current \algname for different schemes.  Likewise, efficient memory allocation is another essential work ahead of this project, where reducing the required hardware enables industry-scale problems to exploit properties in \algname.
	
	\begin{table*}[!ht]
		\caption{Best test accuracy (\%) and its corresponding training time (sec) of \algname, AdaBoost, Random Forest, and Gradient Boosting on binary classification tasks.}
		\centering
		\begin{tabular}{lccccc}
			\toprule
			\textbf{Dataset} & \small\textbf{Criteria} & \small\textbf{\algname} & \small\textbf{AdaBoost} & \small\textbf{Random Forest} & \small\textbf{Gradient Boosting}\\ \midrule\midrule
			\multirow{2}{*}{\small\textbf{Cones}}
			&\small Test	 &$\mathbf{100.0 \pm 0.0}$	 &$\mathbf{100.0 \pm 0.0}$ & $\mathbf{100.0 \pm 0.0}$& $\mathbf{100.0 \pm 0.0}$ \\
			&\small Time	  &$\mathbf{2.81 \pm 0.16}$ & $23.63 \pm 0.14$ & $14.2 \pm 0.06$& $143.87 \pm 1.23$ \\\midrule
			\multirow{2}{*}{\small\textbf{Hill-Valley}}
			&\small Test	 &$\mathbf{99.45 \pm 0.19}$	 &$\mathbf{99.45 \pm 0.19}$ & $99.17 \pm 0.06$& $98.76 \pm 0.34$\\
			&\small Time	 &$\mathbf{0.08 \pm 0.01}$ & $0.37 \pm 0.01$  & $0.14 \pm 0.01$ & $2.92 \pm 0.08$\\\midrule
			\multirow{2}{*}{\small\textbf{Christine}}
			&\small Test	 &$74.27 \pm 0.32$	 &$74.7 \pm 0.01$ & $\mathbf{75.32 \pm 0.41}$ & $76.18 \pm 0.03$ \\
			&\small Time	 &$\mathbf{2.22 \pm 0.2}$ &$92.36 \pm 0.53$  & $6.06 \pm 0.09$ & $108.47 \pm 0.51$ \\\midrule
			\multirow{2}{*}{\small\textbf{Jasmine}}
			&\small Test	 &$80.03 \pm 0.76$ &$\mathbf{81.43 \pm 0.08}$  & $80.7\pm 0.7$ & $80.87 \pm 0.14$ \\
			&\small Time	 &$\mathbf{0.15 \pm 0.04}$ &$2.4 \pm 0.01$  & $0.47 \pm 0.00$ & $2.2 \pm 0.02$ \\\midrule
			\multirow{2}{*}{\small\textbf{Philippine}}
			&\small Test	 &$71.01 \pm 0.37$ &$72.73 \pm 0.01$ & $72.3 \pm 0.16$ & $\mathbf{73.1 \pm 0.47}$ \\
			&\small Time	 &$\mathbf{0.96 \pm 0.08}$ &$11.45 \pm 0.04$  & $5.78 \pm 0.02$ & $74.56 \pm 0.91$ \\\midrule
			\multirow{2}{*}{\small\textbf{SensIT Vehicle}}
			&\small Test	 &$86.23\pm 0.01$ &$85.93 \pm 0.01$  & $86.88 \pm 0.05$ & $\mathbf{87.43 \pm 0.01}$ \\
			&\small Time	 &$\mathbf{25.7\pm 3.03}$ &$576.63 \pm 19.69$  & $303.6 \pm 0.9$ & $730.06 \pm 0.75$ \\\midrule
			\multirow{2}{*}{\small\textbf{Higgs Boson}}
			&\small Test	 &$83.42 \pm 0.06$ &$82.95 \pm 0.01$  & $\mathbf{83.68 \pm 0.04}$ & $82.74 \pm 0.01$ \\
			&\small Time	  &$\mathbf{79.31 \pm 3.55}$ &$270.28 \pm 0.51$  & $258.79 \pm 3.31$ & $397.28 \pm 1.04$ \\\midrule
			\multirow{2}{*}{\small\textbf{MNIST (3,8)}}
			&\small Test	 &$\mathbf{99.31 \pm 0.03}$ &$\mathbf{99.31 \pm 0.02}$  & $98.98 \pm 0.13$ & $98.96 \pm 0.01$ \\
			&\small Time	  &$\mathbf{5.25 \pm 0.76}$ &$81.65 \pm 0.53$  & $6.19 \pm 0.04$ & $99.5 \pm 0.58$ \\\midrule
			\multirow{2}{*}{\small\textbf{MNIST (O,E)}}
			&\small Test	  &$98.15 \pm 0.06$ &$\mathbf{98.17 \pm 0.01}$  & $97.74 \pm 0.04$ & $97.95 \pm 0.03$ \\
			&\small Time	  &$56.34 \pm 0.34$ &$711.22 \pm 3.98$  & $\mathbf{36.62 \pm 0.1}$ & $807.53 \pm 0.67$ \\\midrule
			\multirow{2}{*}{\small\textbf{CIFAR-10 (T,C)}}
			&\small Test	  &$73.67 \pm 0.39$ &$76.12 \pm 0.01$  & $74.24 \pm 0.37$ & $\mathbf{78.5 \pm 0.12}$ \\
			&\small Time	  &$\mathbf{21.44 \pm 4.54}$ &$2282.46$  & $53.44 \pm 0.23$ & $2224.67 \pm 1.63$ \\\midrule
			\multirow{2}{*}{\small\textbf{GAS Drift}}
			&\small Test	  &$99.76 \pm 0.04$ &$\mathbf{99.78 \pm 0.01}$  & $99.66 \pm 0.03$ & $99.68 \pm 0.03 $ \\
			&\small Time	  &$\mathbf{1.13 \pm 0.07}$ &$42.17 \pm 0.03$  & $2.42 \pm 0.03$ & $64.86 \pm 0.1$ \\\midrule
			\multirow{2}{*}{\small\textbf{DNA}}
			&\small Test	  &$97.44 \pm 0.07$ &$97.49 \pm 0.01$ & $\mathbf{97.75 \pm 0.07}$ & $97.65 \pm 0.03$ \\
			&\small Time	  &$\mathbf{0.48 \pm 0.06}$ &$1.84 \pm 0.01$  & $1.08 \pm 0.01$ & $0.97 \pm 0.01$ \\\midrule
			\multirow{2}{*}{\small\textbf{Volkert}}
			&\small Test	  &$74.14\pm 0.08$ &$75.87 \pm 0.3$  & $\mathbf{79.03 \pm 0.16}$ & $77.66 \pm 0.13$ \\
			&\small Time	  &$\mathbf{32.82 \pm 0.97}$ &$541.49 \pm 9.13$  & $74.42 \pm 0.12$ & $553.35 \pm 1.84$ \\
			\bottomrule
		\end{tabular}
		\label{tab:full-comparison}
	\end{table*}
	
	\bibliography{ref}

\begin{thebibliography}{10}

\bibitem{freund1997decision}
Y.~Freund and R.~E. Schapire, ``A decision-theoretic generalization of on-line
  learning and an application to boosting,'' {\em Journal of computer and
  system sciences}, vol.~55, no.~1, pp.~119--139, 1997.

\bibitem{friedman2001greedy}
J.~H. Friedman, ``Greedy function approximation: a gradient boosting machine,''
  {\em Annals of statistics}, pp.~1189--1232, 2001.

\bibitem{lipton2018mythos}
Z.~C. Lipton, ``The mythos of model interpretability,'' {\em Queue}, vol.~16,
  no.~3, pp.~31--57, 2018.

\bibitem{biau2016random}
G.~Biau and E.~Scornet, ``A random forest guided tour,'' {\em Test}, vol.~25,
  no.~2, pp.~197--227, 2016.

\bibitem{benard2020interpretable}
C.~B{\'e}nard, G.~Biau, S.~Da~Veiga, and E.~Scornet, ``Interpretable random
  forests via rule extraction,'' {\em arXiv preprint arXiv:2004.14841}, 2020.

\bibitem{breiman1996bagging}
L.~Breiman, ``Bagging predictors,'' {\em Machine learning}, vol.~24, no.~2,
  pp.~123--140, 1996.

\bibitem{Efron1993AnIT}
B.~Efron and R.~Tibshirani, ``An introduction to the bootstrap,'' 1993.

\bibitem{breiman2001random}
L.~Breiman, ``Random forests,'' {\em Machine learning}, vol.~45, no.~1,
  pp.~5--32, 2001.

\bibitem{louppe2012ensembles}
G.~Louppe and P.~Geurts, ``Ensembles on random patches,'' in {\em Joint
  European Conference on Machine Learning and Knowledge Discovery in
  Databases}, pp.~346--361, Springer, 2012.

\bibitem{srivastava2014dropout}
N.~Srivastava, G.~Hinton, A.~Krizhevsky, I.~Sutskever, and R.~Salakhutdinov,
  ``Dropout: a simple way to prevent neural networks from overfitting,'' {\em
  The journal of machine learning research}, vol.~15, no.~1, pp.~1929--1958,
  2014.

\bibitem{lejeune2019implicit}
D.~LeJeune, H.~Javadi, and R.~G. Baraniuk, ``The implicit regularization of
  ordinary least squares ensembles,'' {\em arXiv preprint arXiv:1910.04743},
  2019.

\bibitem{cortes1995support}
C.~Cortes and V.~Vapnik, ``Support-vector networks,'' {\em Machine learning},
  vol.~20, no.~3, pp.~273--297, 1995.

\bibitem{benard2019sirus}
C.~B{\'e}nard, G.~Biau, S.~Da~Veiga, and E.~Scornet, ``Sirus: making random
  forests interpretable,'' {\em arXiv preprint arXiv:1908.06852}, 2019.

\bibitem{busa2009bandit}
R.~Busa-Fekete and B.~K{\'e}gl, ``Bandit-aided boosting,'' in {\em OPT 2009:
  2nd NIPS Workshop on Optimization for Machine Learning}, 2009.

\bibitem{busa2010fast}
R.~Busa-Fekete and B.~K{\'e}gl, ``Fast boosting using adversarial bandits,'' in
  {\em 27th International Conference on Machine Learning (ICML 2010)},
  pp.~143--150, 2010.

\bibitem{dubout2011tasting}
C.~Dubout and F.~Fleuret, ``Tasting families of features for image
  classification,'' in {\em 2011 International Conference on Computer Vision},
  pp.~929--936, IEEE, 2011.

\bibitem{dubout2011boosting}
C.~Dubout and F.~Fleuret, ``Boosting with maximum adaptive sampling,'' in {\em
  Advances in Neural Information Processing Systems}, pp.~1332--1340, 2011.

\bibitem{dubout2014adaptive}
C.~Dubout and F.~Fleuret, ``Adaptive sampling for large scale boosting,'' {\em
  The Journal of Machine Learning Research}, vol.~15, no.~1, pp.~1431--1453,
  2014.

\bibitem{domingo2000madaboost}
C.~Domingo, O.~Watanabe, {\em et~al.}, ``Madaboost: A modification of
  adaboost,'' in {\em COLT}, pp.~180--189, Citeseer, 2000.

\bibitem{bradley2008filterboost}
J.~K. Bradley and R.~E. Schapire, ``Filterboost: Regression and classification
  on large datasets,'' in {\em Advances in neural information processing
  systems}, pp.~185--192, 2008.

\bibitem{friedman2002stochastic}
J.~H. Friedman, ``Stochastic gradient boosting,'' {\em Computational statistics
  \& data analysis}, vol.~38, no.~4, pp.~367--378, 2002.

\bibitem{chen2016xgboost}
T.~Chen and C.~Guestrin, ``Xgboost: A scalable tree boosting system,'' in {\em
  Proceedings of the 22nd acm sigkdd international conference on knowledge
  discovery and data mining}, pp.~785--794, ACM, 2016.

\bibitem{ke2017lightgbm}
G.~Ke, Q.~Meng, T.~Finley, T.~Wang, W.~Chen, W.~Ma, Q.~Ye, and T.-Y. Liu,
  ``Lightgbm: A highly efficient gradient boosting decision tree,'' in {\em
  Advances in neural information processing systems}, pp.~3146--3154, 2017.

\bibitem{dorogush2018catboost}
A.~V. Dorogush, V.~Ershov, and A.~Gulin, ``Catboost: gradient boosting with
  categorical features support,'' {\em arXiv preprint arXiv:1810.11363}, 2018.

\bibitem{ibragimov2019minimal}
B.~Ibragimov and G.~Gusev, ``Minimal variance sampling in stochastic gradient
  boosting,'' in {\em Advances in Neural Information Processing Systems},
  pp.~15061--15071, 2019.

\bibitem{mohri2018foundations}
M.~Mohri, A.~Rostamizadeh, and A.~Talwalkar, {\em Foundations of machine
  learning}.
\newblock MIT press, 2018.

\bibitem{breiman1984classification}
L.~Breiman, J.~Friedman, C.~J. Stone, and R.~A. Olshen, {\em Classification and
  regression trees}.
\newblock CRC press, 1984.

\bibitem{wyner2017explaining}
A.~J. Wyner, M.~Olson, J.~Bleich, and D.~Mease, ``Explaining the success of
  adaboost and random forests as interpolating classifiers,'' {\em The Journal
  of Machine Learning Research}, vol.~18, no.~1, pp.~1558--1590, 2017.

\bibitem{friedman2000additive}
J.~Friedman, T.~Hastie, R.~Tibshirani, {\em et~al.}, ``Additive logistic
  regression: a statistical view of boosting (with discussion and a rejoinder
  by the authors),'' {\em The annals of statistics}, vol.~28, no.~2,
  pp.~337--407, 2000.

\bibitem{altmann2010permutation}
A.~Altmann, L.~Tolo{\c{s}}i, O.~Sander, and T.~Lengauer, ``Permutation
  importance: a corrected feature importance measure,'' {\em Bioinformatics},
  vol.~26, no.~10, pp.~1340--1347, 2010.

\bibitem{nembrini2018revival}
S.~Nembrini, I.~R. K{\"o}nig, and M.~N. Wright, ``The revival of the gini
  importance?,'' {\em Bioinformatics}, vol.~34, no.~21, pp.~3711--3718, 2018.

\bibitem{Meijer2016RegularizingAW}
D.~W.~J. Meijer and D.~M.~J. Tax, ``Regularizing adaboost with validation sets
  of increasing size,'' {\em 2016 23rd International Conference on Pattern
  Recognition (ICPR)}, pp.~192--197, 2016.

\bibitem{breiman1996out}
L.~Breiman, ``Out-of-bag estimation,'' 1996.

\bibitem{schapire1997using}
R.~E. Schapire, ``Using output codes to boost multiclass learning problems,''
  in {\em ICML}, vol.~97, pp.~313--321, Citeseer, 1997.

\bibitem{chengsheng2017adaboost}
T.~Chengsheng, L.~Huacheng, and X.~Bing, ``Adaboost typical algorithm and its
  application research,'' in {\em MATEC Web of Conferences}, vol.~139,
  p.~00222, EDP Sciences, 2017.

\bibitem{drucker1997improving}
H.~Drucker, ``Improving regressors using boosting techniques,'' in {\em ICML},
  vol.~97, pp.~107--115, 1997.

\bibitem{solomatine2004adaboost}
D.~P. Solomatine and D.~L. Shrestha, ``Adaboost. rt: a boosting algorithm for
  regression problems,'' in {\em 2004 IEEE International Joint Conference on
  Neural Networks (IEEE Cat. No. 04CH37541)}, vol.~2, pp.~1163--1168, IEEE,
  2004.

\bibitem{lecun1998mnist}
Y.~LeCun, ``The mnist database of handwritten digits,'' {\em http://yann.
  lecun. com/exdb/mnist/}, 1998.

\bibitem{Dua:2019}
D.~Dua and C.~Graff, ``{UCI} machine learning repository,'' 2017.

\bibitem{krizhevsky2009learning}
A.~Krizhevsky, G.~Hinton, {\em et~al.}, ``Learning multiple layers of features
  from tiny images,'' 2009.

\bibitem{scikit-learn}
F.~Pedregosa, G.~Varoquaux, A.~Gramfort, V.~Michel, B.~Thirion, O.~Grisel,
  M.~Blondel, P.~Prettenhofer, R.~Weiss, V.~Dubourg, J.~Vanderplas, A.~Passos,
  D.~Cournapeau, M.~Brucher, M.~Perrot, and E.~Duchesnay, ``Scikit-learn:
  Machine learning in {P}ython,'' {\em Journal of Machine Learning Research},
  vol.~12, pp.~2825--2830, 2011.

\end{thebibliography}
	\bibliographystyle{ieeetr}
	
	\appendices
	\section{Dataset}\label{sec:datasets}
	According to our specific problem, we select datasets listed in Table~\ref{tab:dataset} from UCI machine learning repository~\cite{Dua:2019}, MNIST~\cite{lecun1998mnist}, and CIFAR~\cite{krizhevsky2009learning}. Note that we consider multiclass datasets like MNIST and CIFAR-10 and select two classes that are harder to be distinguished like digits $3 \& 8$ from MNIST or "Truck \& Car" classes from CIFAR-10. Additionally, we partition all classes into two for different multiclass datasets, like "Odd \& Even" digits in MNIST.  About 20\% of each dataset is selected as the test data. "Cones" is a sparse synthetic dataset of two originally $10$-dimensional cones. Adding $490$ noise features to them, we transform data points to $\bbR^{500}$.
	\begin{table}[H]
		\centering
		\caption{Size of the datasets in Tables~\ref{tab:fixed-runtime} and~\ref{tab:full-comparison}}
		\begin{tabular}{llll}
			\toprule
			Dataset                  & $N_\text{train}$      & $M$   &  $N_\text{test}$   \\
			\cmidrule(lr){2-3}
			Cones                    &  20000    &   500   &   5000\\
			Hill-Valley              &  1000     &   100   &   200\\
			Christine                &  4300     &   1636  &   1100\\
			Jasmine                  &  2400     &   144   &   600\\
			Philippine               &  4700     &   308   &   1100\\
			SensIT Vehicle           &  78800    &   100   &   19700\\
			Higgs Boson              &  200000   &   30    &   50000\\
			MNIST (3 vs. 8)          &  12000    &   784   &   2000\\
			MNIST (Odd vs. Even)     &  60000    &   784   &   10000\\
			CIFAR-10 (Truck vs. Car) &  10000    &   3072  &   2000\\
			Gas Drift                &  11100    &   128   &   2800\\
			DNA                      &  2600     &   180   &   6000\\
			Volkert                  &  46600    &   180   &   11700\\
			Fabert                   &  6600     &   800   &   1600\\
			\bottomrule
		\end{tabular}
		\label{tab:dataset}
	\end{table}
	
	\section{Stopping Criterion Algorithm}\label{sec:stopping-criterion}
	The trend in $\oop$ is an approximation of the generalization accuracy; however, there are numerical oscillations in it.  To make the stopping algorithm (Algorithm~\ref{alg:stop}) robust to such numerical vibrations, instead of the largest previous value of $\oop$, it keeps the $k$ highest previous values of $\oop$ in a list, $\mcA$, and compares the $\oop^\ptp$ with their minimum.  If the $\oop$ value does not improve after $k$ sequential iterations, then \algname halts.
	\begin{algorithm}[H]
		\caption{Stopping-Criterion}
		$\gamma = 1 + \frac{\log(n)}{N}$ \hspace{1em}\textcolor{commentcolor}{// ratio of the growth must be $\gamma>1$}\\
		$k = \log(N)$ \hspace{2em}\textcolor{commentcolor}{// number of top $\oop$}\\
		$v = 0$ \hspace{1em}\textcolor{commentcolor}{// number of successive iterations without progress}\\
		$\mcA=$ a list containing $k$ zeros \hspace{1em}\textcolor{commentcolor}{// keeping the top $k$ largest values of $\oop$ up to each iteration}\\
		\textbf{Stopping-Criterion} $(\oop^\ptp)$
		\begin{enumerate}
			\item \textcolor{blue}{\textbf{Detect the best iteration:}}\label{step:detect-best-iteration}
			\begin{enumerate}
				\item[] if $\oop^\ptp > \max(\mcA)$, set $T=t$ 
			\end{enumerate}
			\item \textcolor{blue}{\textbf{Check the stopping criteria:}}\label{step:check-stop}
			\begin{enumerate}
				\item[] if $v>k$ or $t>T_{\max}$, halt Algorithm~\ref{alg:main}\\
				\phantom{0}\hspace{3em}\textcolor{commentcolor}{// $T_{\max}:$ maximum number of iterations}
			\end{enumerate}
			\item \textcolor{blue}{\textbf{Check the progress of oop:}}\label{step:check-progress}
			\begin{enumerate}
				\item[] $v = \begin{cases}
				v+1 & \oop^\ptp< \gamma.\min(\mcA)\\
				0 & \text{otherwise}
				\end{cases}$

			\end{enumerate}
			\item \textcolor{blue}{\textbf{Update $\mcA$:}}\label{step:update-best-accuracies-list}
			\begin{enumerate}
				\item[] if $\oop^\ptp>\min(\mcA)$, then replace the minimum entry with $\oop^\ptp$

			\end{enumerate}
		\end{enumerate}
		\label{alg:stop}
	\end{algorithm}

\end{document}